\documentclass[10pt,twocolumn,letterpaper]{article}

\usepackage{cvpr}              %

\definecolor{cvprblue}{rgb}{0.21,0.49,0.74}
\usepackage[pagebackref,breaklinks,colorlinks,allcolors=cvprblue]{hyperref}
\usepackage{multirow} 
\usepackage{amsmath}
\usepackage{amssymb}
\def\paperID{4713} %
\def\confName{CVPR}
\def\confYear{2025}

\title{FitDiT: Advancing the Authentic Garment Details for High-fidelity \\ Virtual Try-on}

\author{
\textbf{Boyuan Jiang}\textsuperscript{1*}
\ 
\textbf{Xiaobin Hu}\textsuperscript{1*}
\ 
\textbf{Donghao Luo}\textsuperscript{1}
\ 
\textbf{Qingdong He}\textsuperscript{1}
\ 
\textbf{Chengming Xu}\textsuperscript{1} 
\\
\textbf{Jinlong Peng}\textsuperscript{1} 
\
\textbf{Jiangning Zhang}\textsuperscript{1} 
\ 
\textbf{Chengjie Wang}\textsuperscript{1} 
\ 
\textbf{Yunsheng Wu}\textsuperscript{1} 
\ 
\textbf{Yanwei Fu}\textsuperscript{2} 
\\
\textsuperscript{1} Tencent
\quad
\textsuperscript{2} Fudan University
}

\begin{document}

\twocolumn[{%
\renewcommand\twocolumn[1][]{#1}%
\maketitle
\begin{center}
    \centering
    \captionsetup{type=figure}
\includegraphics[width=1\textwidth]{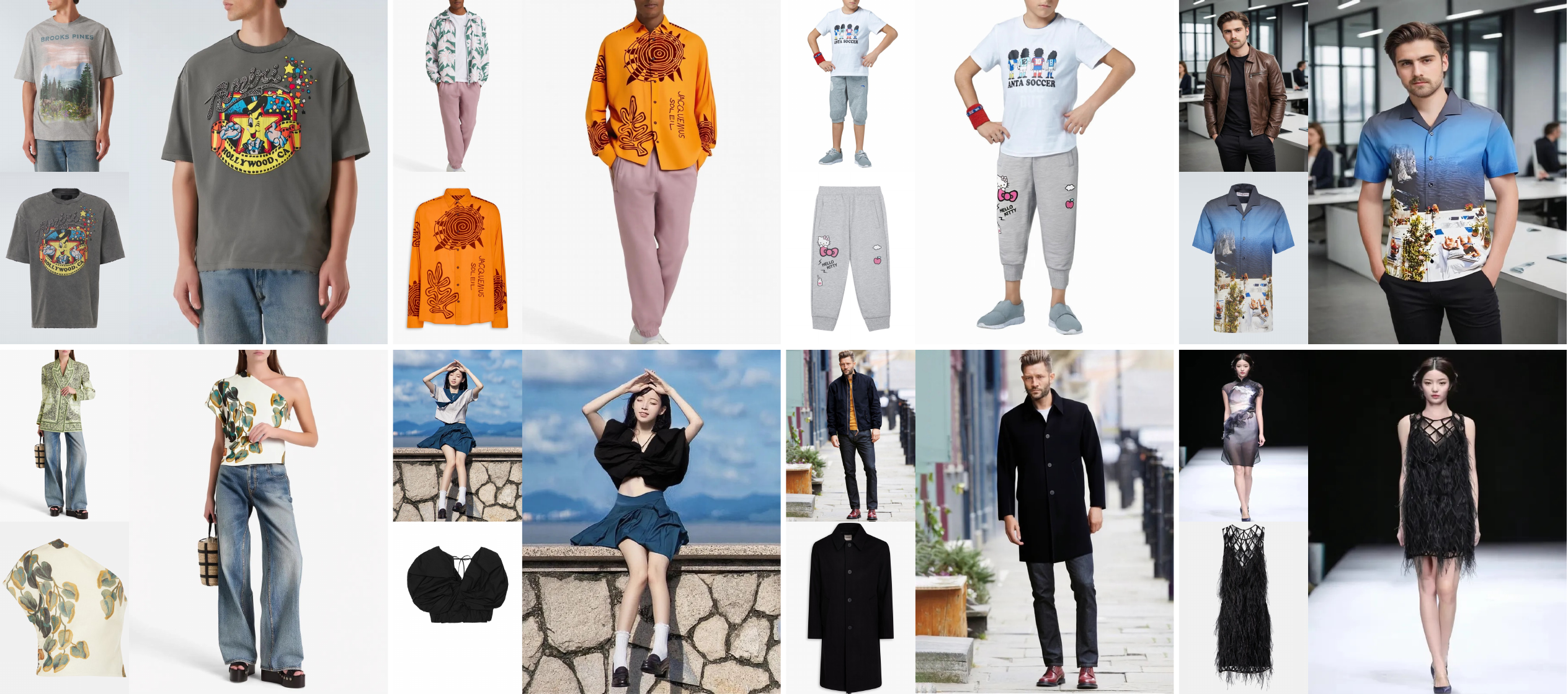}
    \vspace{-10pt}
    \captionof{figure}{\small FitDiT demonstrates exceptional performance in virtual try-on, addressing challenges related to texture-aware preservation and size-aware fitting across various scenarios.
    }
    \label{fig:abs}
\end{center}%
}]

\if TT\insert\footins{\noindent\footnotesize{
*Equal Contribution  \\
Project page: \url{https://byjiang.com/FitDiT/}.}}\fi

\begin{abstract}
Although image-based virtual try-on has made considerable progress, emerging approaches still encounter challenges in producing high-fidelity and robust fitting images across diverse scenarios. These methods often struggle with issues such as texture-aware maintenance and size-aware fitting, which hinder their overall effectiveness.
To address these limitations, we propose a novel garment perception enhancement technique, termed FitDiT, designed for high-fidelity virtual try-on using Diffusion Transformers (DiT) allocating more parameters and attention to high-resolution features.
First, to further improve texture-aware maintenance, we introduce a garment texture extractor that incorporates garment priors evolution to fine-tune garment feature, facilitating to better capture rich details such as stripes, patterns, and text.
Additionally, we introduce frequency-domain learning by customizing a frequency distance loss to enhance high-frequency garment details.
To tackle the size-aware fitting issue, we employ a dilated-relaxed mask strategy that adapts to the correct length of garments, preventing the generation of garments that fill the entire mask area during cross-category try-on.
Equipped with the above design, FitDiT surpasses all baselines in both qualitative and quantitative evaluations. It excels in producing well-fitting garments with photorealistic and intricate details, while also achieving competitive inference times of 4.57 seconds for a single $1024\times 768$ image after DiT structure slimming, outperforming existing methods. The code and dataset will be made publicly available.

\end{abstract}
    
\section{Introduction}
\label{sec:intro}
The remarkable growth of e-commerce has created a consistent demand for a more convenient and personalized shopping experience. Image-based virtual try-on (VTON) has emerged as a widely adopted technique aimed at generating realistic images of human models wearing specific garments, thereby enhancing the shopping experience for consumers. In recent years, a multitude of researchers \cite{dong2019towards, ge2021disentangled, issenhuth2020not, han2019clothflow, han2018viton, he2022style, kim2024stableviton, minar2020cp, wang2018toward, xie2023gp, yang2020towards} have dedicated significant efforts towards achieving more realistic and photorealistic virtual try-on results.

Most recent studies in this domain have predominantly utilized generative adversarial networks (GANs) \cite{goodfellow2020generative} or latent diffusion models (LDMs) \cite{rombach2022high} for image and video synthesis. However, traditional GAN-based approaches \cite{han2019clothflow, han2018viton, he2022style} often struggle to faithfully reproduce complex garment textures, realistic lighting and shadows, and lifelike depictions of the human body. As a result, recent research has shifted focus toward LDM U-Net-based methods \cite{zhu2023tryondiffusion, xu2024ootdiffusion, choi2024improving}, which enhance the authenticity of clothed images. Despite these advancements, virtual try-on still faces significant garment-fitting challenges, particularly in two areas: (1) \textbf{Rich texture-aware maintenance}, where the transformation of intricate textures (e.g., patterns, text, stripes, trademarks) to the target model is hindered by the limitations of U-Net-based diffusion structures, which allocate less attention to high-resolution latent features; and (2) \textbf{Size-aware fitting}, where clothing information leakage occurs in cross-category or size-mismatched try-on scenarios, leading to generated garments that cover the entire agnostic mask region. To address these challenges and ensure robustness across diverse scenarios, we propose FitDiT, a novel approach to enhance authentic garment perception for high-fidelity virtual try-on.

First, to tackle the rich texture-aware challenge, we introduce a DiT-based LDM that leverages the strengths of the Diffusion Transformer architecture, focusing more attention on high-resolution latent features related to garment patterns. We also provide an analysis of the structural differences between U-Net and DiT-based LDMs.
Given that existing DiT methods currently available suffer from redundancy and sub-optimal structures for virtual try-on applications, we propose a reevaluation of the try-on DiT approach. The customization design of our FitDiT mainly focuses on three key aspects: structure slimming, garment condition modulation, and garment feature injection.
Besides, unlike other try-on methods that directly use existing Diffusion U-Net as the Garment Net without adequately fitting the garment texture, we introduce a garment priors evolution strategy. This strategy fine-tunes the garment-dedicated extractor using garment data to render exceptional and rich details.
Furthermore, from the observation of the frequency spectra discrepancies between synthesized and real images, a frequency-spectra distance loss in pixel space is proposed to refine the high-frequency knowledge in the frequency domain. 
This high-frequency knowledge involves complex textures and boosts authentic garment perception. 
Lastly, to improve size-aware fitting, we propose a dilated-relaxed mask augmentation that employs a coarse rectangular mask with random adjustments to its length and width. This masking strategy can effectively avoid the leakage of garment shape information during training. The random adjustments enable the model to autonomously perceive the overall shape of the garment, promoting the generation of well-fitted try-ons, particularly for mismatched garments. Our contributions are summarized as follows:
\begin{itemize}[topsep=3pt, partopsep=3pt,leftmargin=5pt, itemsep=3pt]
\item To the best of our knowledge, our FitDiT is the first attempt to customize the Diffusion Transformer (DiT) for virtual try-on applications, overcoming the limitations of current U-Net LDMs in complex texture maintenance by assigning greater attention to high-resolution features.
\item  For rich texture-aware maintenance, we propose a garment priors evolution strategy to better exact the pattern knowledge of garments and a frequency-spectra distance loss in pixel space to retain complex patterns.
\item For size-aware fitting, we propose a dilated-relaxed mask augmentation with the coarse rectangular mask to lower the leakage of garment shape, and enable the model to adaptively learn the overall shape of garments.
\item Extensive qualitative and quantitative evaluations have clearly demonstrated FitDiT's superiority over state-of-the-art virtual try-on models, especially in handling richly textured garments with size mismatches. Additionally, it achieves competitive inference times of 4.57 seconds for a single $1024 \times 768$ image, surpassing existing methods. These findings serve as a significant milestone in advancing the field of virtual try-on, enabling more intricate applications in real-world settings.
\end{itemize}

\begin{figure*}[htb]
    \centering
    \includegraphics[width=0.9\textwidth]{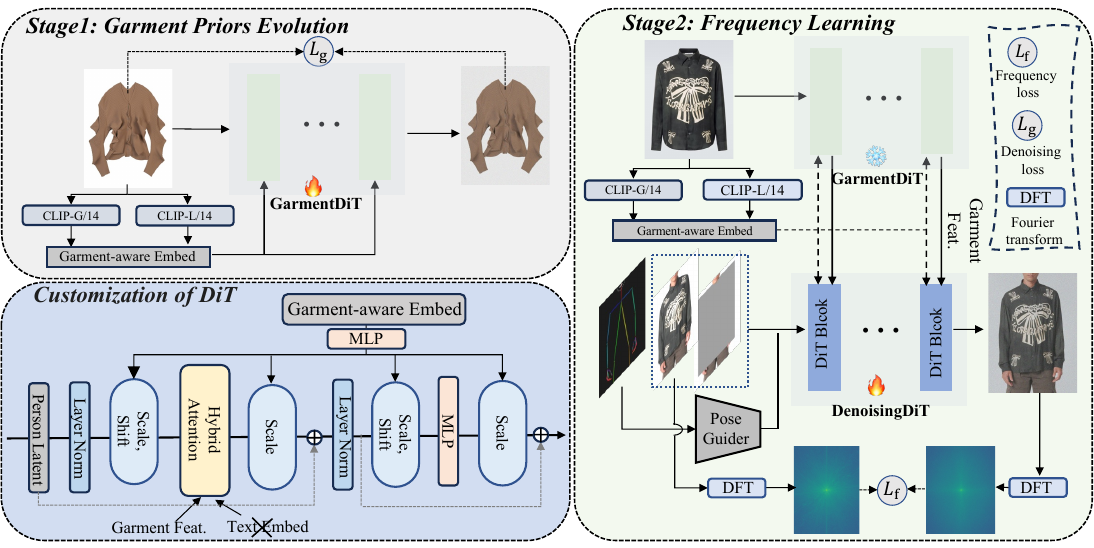}
    \caption{FitDiT employs a two-stage training strategy. In the first stage, Garment Priors Evolution is utilized to fine-tune GarmentDiT for enhanced clothing feature extraction. In the second stage, we customize the DiT blocks through structure slimming, garment condition modulation, and high-resolution garment feature injection, resulting in DenoisingDiT for the try-on. DenoisingDiT is trained jointly using frequency loss and denoising loss.
    }
    \label{fig:pipeline}
    \vspace{-5mm}
\end{figure*}

\section{Related Works}
\subsection{Image-based Virtual Try-on}
Image-based virtual try-on has been extensively researched over the years, emerging as a promising and formidable task. 
A series of studies based on the Generative Adversarial Networks (GANs) \cite{lee2022high, men2020controllable,xie2023gp,yang2023occlumix} have been carried out for more natural generation. 
However, GAN-based methods often struggle to generate outfitted images in high-fidelity and realism due to the fact that the GAN-based framework uses numerous efforts for explicit the warping process and overlooks realistic garment texture. Meanwhile, GAN-based methods \cite{ge2021parser,issenhuth2020not,lee2022high} lack the ability to generalize well to diverse person images, leading to significant performance degradation when applied to out-of-distribution images. As remarkable advancements have been observed in Text-to-Image diffusion models \cite{saharia2022photorealistic, ruiz2023dreambooth, hu2024diffumatting} in recent years, some studies \cite{chen2024wear, liang2024vton, kolors} have been encouraged to integrate pre-trained diffusion models \cite{rombach2022high} as generative priors into the virtual try-on task. 
For instance, TryOnDiffusion \cite{zhu2023tryondiffusion} introduces parallel U-Nets Diffusion to enhance garment details and warp the garment for try-on. 
Subsequent studies have regarded virtual try-on as an exemplar-based image inpainting problem, focusing on fine-tuning inpainting diffusion models using virtual try-on datasets to produce high-quality try-on images.
For example, LADI-VTON \cite{morelli2023ladi}and DCI-VTON \cite{gou2023taming} have been proposed that treat clothing as pseudo-words or utilize warping networks to seamlessly integrate garments into pre-trained diffusion models.
IDM-VTON \cite{choi2024improving} designs advanced attention modules to encode high-level garment semantics while extracting low-level features to preserve fine-grained details. Anyfit \cite{li2024anyfit} employs a diffusion model and introduces a Hydra Block for attire combinations, producing harmonized upper and lower styles. Although these methods have garnered attention regarding the nature and realism of synthesized images, they still face challenges in addressing the hard occasions of rich texture-aware maintenance and size-aware fitting, which are prevalent in real-world virtual try-on.

\subsection{Diffusion Models}
U-Net \cite{ronneberger2015u} has been the established backbone for diffusion-based image generation and is extensively employed in text-to-image models such as LDM \cite{rombach2022high}, SDXL \cite{podell2023sdxl}, DALL-E \cite{betker2023improving}, and Imagen \cite{saharia2022photorealistic}. 
U-ViT \cite{bao2023all} utilizes ViT with long skip connections between shallow and deep layers, highlighting the significance of these long skip connections while suggesting that the downsampling and upsampling operators in CNN-based U-Net are not always essential. 
DiT \cite{peebles2023scalable} uses the Transformers to replace U-Net for class-conditioned image generation and proves the significant correlation existing between network complexity and sample quality.
Stable Diffusion 3 \cite{esser2024scaling} and Flux introduces a transformer-based architecture that utilizes separate weights for text and image modalities, achieving an amazing text-to-image effect.
In this paper, we introduce the DiT structure and analyze its advantages for virtual try-on, highlighting the limitations of U-Net-based LDMs. The downsampling and upsampling operations in U-Net are not essential for our purposes, as they tend to focus more attention on low-resolution latent features. This focus undermines the maintenance of high-resolution, rich textures in garments, which is crucial for achieving realistic virtual try-on.

\section{Method}
\subsection{Model Overview}
An overall framework of FitDiT is presented in Fig.~\ref{fig:pipeline}. Given a person image $x_p$ and a garment image $x_g$, FitDiT aims to generate an image $x_{tr}$ that visualizes the person wearing the provided garment. It is common practice to treat virtual try-on as a specific case of exemplar-based image inpainting task, which involves filling the masked person image $x_p$ using the garment $x_g$ as a reference. FitDiT employs a parallel-branch architecture, where the GarmentDiT extracts detailed garment features from the input garment image. These features are then injected into DenoisingDiT through a hybrid attention mechanism. We utilize a customized Stable Diffusion 3~\cite{esser2024scaling} for both GarmentDiT and DenoisingDiT, which will be described in detail in Sec.~\ref{Customization}. Additionally, we incorporate a Pose Guider~\cite{hu2024animate}, composed of a 4-layer convolutional network (with 4 × 4 kernels, 2 × 2 strides, and 32, 64, 256, and 512 channels), which uses DWPose~\cite{yang2023effective} as input to ensure the coherence of the generated human body in the inpainting area.

\subsection{Customization of DiT for Virtual Try-on}
\label{Customization}
The original SD3 is a text-to-image model composed of a series of stacked MM-DiT blocks. We analyze the differences between the text-to-image and virtual try-on tasks and customize the model specifically for the virtual try-on.

\noindent\textbf{Structure slimming.} The original SD3 uses OpenCLIP bigG/14~\cite{cherti2023reproducible}, CLIP-ViT/L~\cite{radford2021learning} and T5-xxl~\cite{raffel2020exploring} as text encoders to process the text prompt, which serves as a control condition for image generation. However, for virtual try-on, the generated image is primarily determined by the given garment, with the text prompt having limited impact~\cite{chong2024catvton}.
Therefore, we remove the text encoder from SD3, resulting in approximately 72\% parameter savings. This modification also improves the speed of model training and inference while reducing memory usage.

\noindent\textbf{Garment condition modulation.} In virtual try-on tasks, various types of garments (\textit{e.g.,} upper body, lower body, dresses) are typically trained using a unified model, which may lead to confusion during training.  OOTDiffusion~\cite{xu2024ootdiffusion} utilizes a text embedding of the garment label $y \in \{"upperbody", "lowerbody", "dress"\}$ as an auxiliary conditioning input to differentiate between garment types. However, this condition is considered coarse-grained, as garments within the same category can vary significantly. To this end, we propose utilizing the image encoders of OpenCLIP bigG/14 and CLIP-ViT/L to encode the given garment into a garment image embedding. This embedding is then combined with the timestep embedding to produce scale and shift parameters that modulate features in the DiT block in a garment-aware manner.

\noindent\textbf{Garment feature injection.} To extract garment features, we first input the garment into the GarmentDiT and forward it with timestep $t=0$, saving the key and value features $\{k_r, v_r\}$ from the GarmentDiT attention module, which contains rich clothing texture information. Then during each denoising step, we inject the saved garment features into DenoisingDiT using the hybrid attention mechanism with the following q, k, v,
\begin{equation}
\begin{aligned}
\label{eq:hybrid_atten}
    q=q_{d}, k=k_{d}\copyright k_{r}, v=v_{d}\copyright v_{r},
\end{aligned}
\end{equation}
where $\copyright$ denotes feature concatenation across the channel dimension, and $q_d$, $k_d$, $v_d$ represent the query, key, and value from DenoisingDiT, respectively.

\subsection{Dilated-relaxed Mask Strategy}
\begin{figure}[htb]
    \centering
    \includegraphics[width=0.98\linewidth]{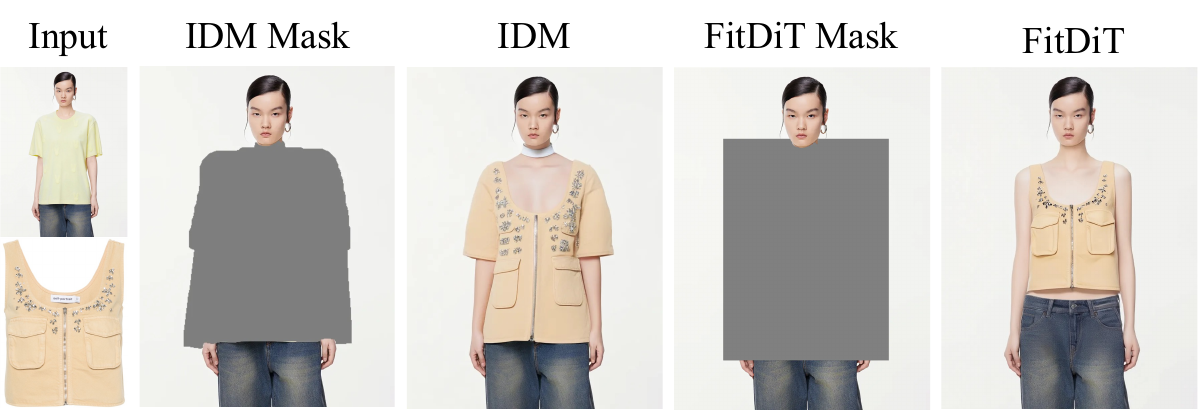}
    \vspace{-3mm}
    \caption{Previous works tend to fill the entire inpainting area due to a strict mask strategy. In contrast, FitDiT can accurately restore the shape of the garment with the dilated-relaxed mask strategy.}
    \label{fig:relaxed_mask}
    \vspace{-5mm}
\end{figure}
Existing works on cross-category try-on often encounters inaccuracies in shape rendering. This issue arises because previous approaches typically construct agnostic masks strictly based on human parsing contours, as illustrated in Fig.~\ref{fig:relaxed_mask}. Such a mask construction strategy can lead to the leakage of garment shape information during training, resulting in the model's tendency to fill the entire mask area during inference.

To mitigate this problem, we propose a dilated-relaxed mask strategy that allows the model to automatically learn the optimal length of the target garment during training. Specifically, for upper-body try-on, we derive the smallest enclosing rectangle that can simultaneously encompass the hand keypoints and the human parsing of the upper garment. For lower-body try-on, we construct agnostic masks using foot keypoints and human parsing of the lower garment. For dresses, we consider both hand and foot keypoints, along with human parsing of the dresses. Furthermore, to enhance the model's perception of garment shapes, we randomly expand the edges of the masks by a few pixels to cover parts of the non-changing areas.

\subsection{Garment Texture Enhancement}
To maintain rich texture during try-on, we propose a two-stage training strategy. First, we conduct a garment priors evolution stage to fine-tune the GarmentDiT using rich garment data, enabling it to render with exceptional detail. This is followed by DenoisingDiT training, which incorporates frequency loss and denoising loss.
\paragraph{Garment priors evolution}
The garment feature extractor plays a crucial role in preserving texture details during the try-on task. In previous works~\cite{xu2024ootdiffusion, choi2024improving}, its weights are initialized using models trained on large-scale image-text paired datasets without fine-tuning. 
However, due to the discrepancy between text-to-image and try-on tasks, directly applying text-to-image models as feature extractors for the try-on task may lead to suboptimal performance. To address this issue, we propose a simple yet effective garment priors evolution strategy to enhance our GarmentDiT. Specifically, given a garment latent $z_0$, we fine-tune the GarmentDiT using the following denoising loss function:
\begin{equation}
\begin{aligned}
\label{eq:garment_loss}
    L_{g}&=\mathbb{E}_{\epsilon \sim \mathcal{N}(0,1), t \sim \mathcal{U}(t)}[w(t)||\epsilon_\theta(z_{t};I_{vec},t)-\epsilon||^2],
\end{aligned}
\end{equation}
where $z_{t}=(1-t)z_0+t\epsilon$ is the noisy garment latent, and $w(t)$ is a weighting function at each timestep $t$. The model trained with $L_{g}$ loss will serve as the garment feature extractor and will remain frozen during subsequent training.

\begin{figure}[tb]
    \centering
    \includegraphics[width=0.8\linewidth]{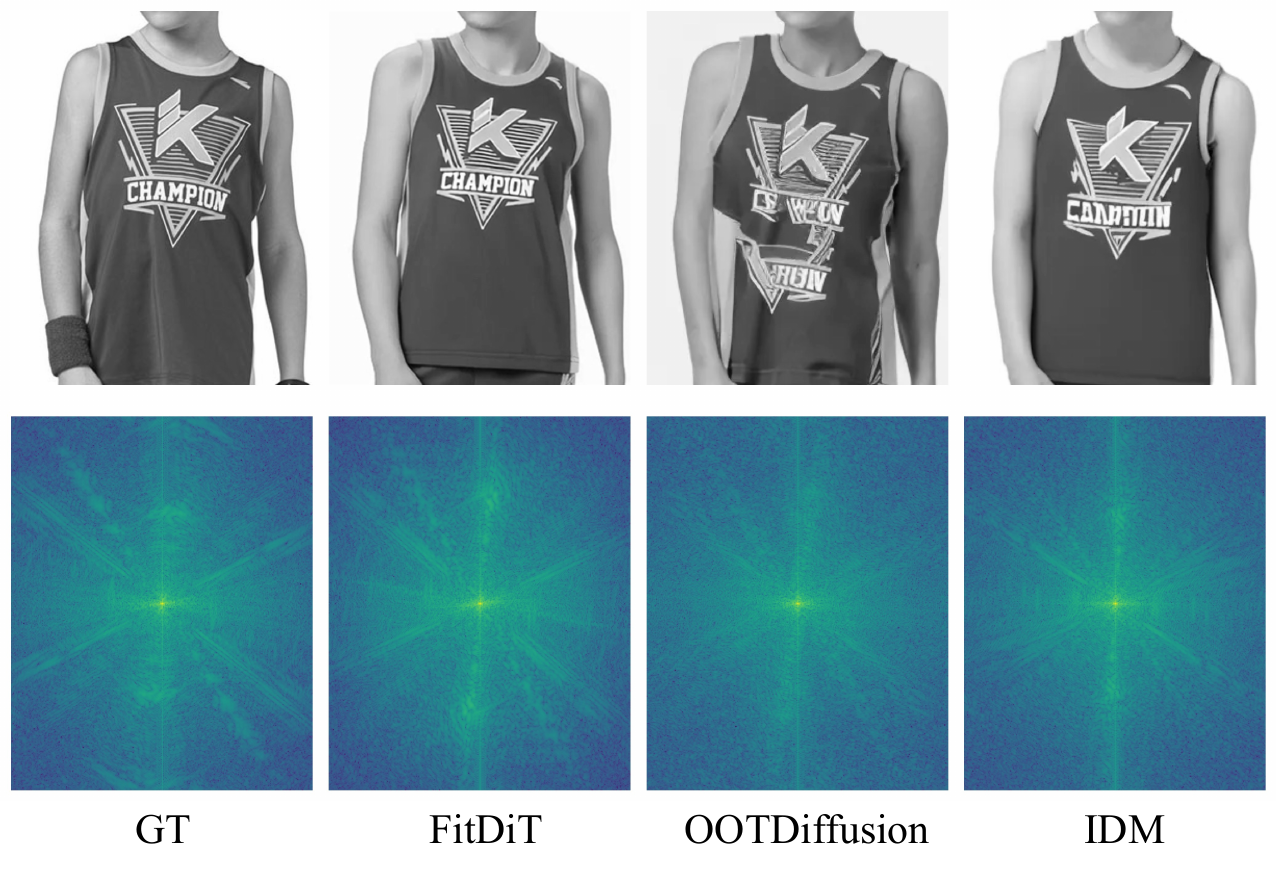}
    \vspace{-3mm}
    \caption{Frequency domain gaps between the real and the generated images by different algorithms.}
    \label{fig:freq}
    \vspace{-5mm}
\end{figure}
\paragraph{Frequency learning}
The Discrete Fourier Transform (DFT) is a mathematical method that changes a discrete, finite signal into its frequency components, which are shown as complex exponential waves. An image is a type of two-dimensional discrete signal made up of real numbers. In Fig.~\ref{fig:freq}, we visualize the spectrum of a real image alongside the try-on results generated by different algorithms using DFT. It is evident that significant discrepancies in the details of the generated images, such as text and patterns, compared to the real images result in noticeable gaps in their spectral representations. This observation leads us to hypothesize that minimizing the differences between the generated images and the real images in the frequency domain could enhance the fidelity of clothing detail reconstruction in the try-on results. To this end, we propose a frequency-spectra distance loss in pixel space, which enables the model to focus more on the components with significant gaps in the frequency domain during optimization.
To represent an image in the frequency domain, we use the 2D DFT:
\begin{align}
    F(u,v)&= \sum_{x=0}^{M-1}\sum_{y=0}^{N-1}f(x,y)\times e^{-i2\pi(\frac{ux}{M}+\frac{vy}{N})}, \label{eq:DFT} \\
    e^{-i2\pi(\frac{ux}{M}+\frac{vy}{N})}&=\cos{2\pi(\frac{ux}{M}+\frac{vy}{N})}-i\sin{2\pi(\frac{ux}{M}+\frac{vy}{N})}, \notag
\end{align}
where $M\times N$ is the image size, $f(x,y)$ represents the pixel at coordinate $(x,y)$ in the spatial domain, and $F(u,v)$ is the complex frequency value that corresponds to a spatial frequency at coordinate $(u,v)$ in the frequency spectrum. $e$ and $i$ are Euler’s number and the imaginary unit. 
Different frequencies in the frequency spectrum correspond to various patterns within the image. Therefore, by minimizing the spectral error, we can enhance the overall structural and textural accuracy of the generated results.

Specifically, to compute the frequency domain error, we first need to convert the noise latent $z_t$ back to pixel space. Following the forward process of Rectified Flows (RFs)~\cite{liu2022flow}, the model input $z_t$ at the current timestep $t$ is a straight path between the data distribution $z_0$ and a standard normal distribution $\epsilon \sim \mathcal{N}(0,1)$, i.e.,
\begin{equation}
\begin{aligned}
\label{eq:forward}
    z_{t}&=  (1-t)z_0+t\epsilon.
\end{aligned}
\end{equation}
Based on Eq.~\ref{eq:forward}, we can estimate $\hat{z}_0$ through a single denoising step,  utilizing the noise $\epsilon_t$ predicted by the network at the current timestep $t$,
\begin{equation}
\begin{aligned}
    \hat{z}_0&= \frac{z_t-t\epsilon_t}{1-t}.
\end{aligned}
\end{equation}
The latent $\hat{z}_0$ is converted back to pixel space by the VAE decoder, resulting in the predicted try-on image $\hat{x}_{tr}$. We can calculate the frequency gap between $\hat{x}_{tr}$ and $x_{p}$,
\begin{equation}
\begin{aligned}
\label{eq:freq_loss}
    L_{f}&= \sum_{u=0}^{M-1}\sum_{v=0}^{N-1}|F_{\hat{x}_{tr}\odot m_g}(u,v)-F_{x_{p}\odot m_g}(u,v)|^2,
\end{aligned}
\end{equation}
where $m_g$ represents the garment segmentation mask of $x_{p}$ obtained from human parsing model. Ultimately, the overall training loss function of DenoisingDiT is a combination of the imaged-based frequency loss and the latent-based diffusion noise prediction loss.

\begin{figure}[tb]
    \centering
    \includegraphics[width=0.75\linewidth]{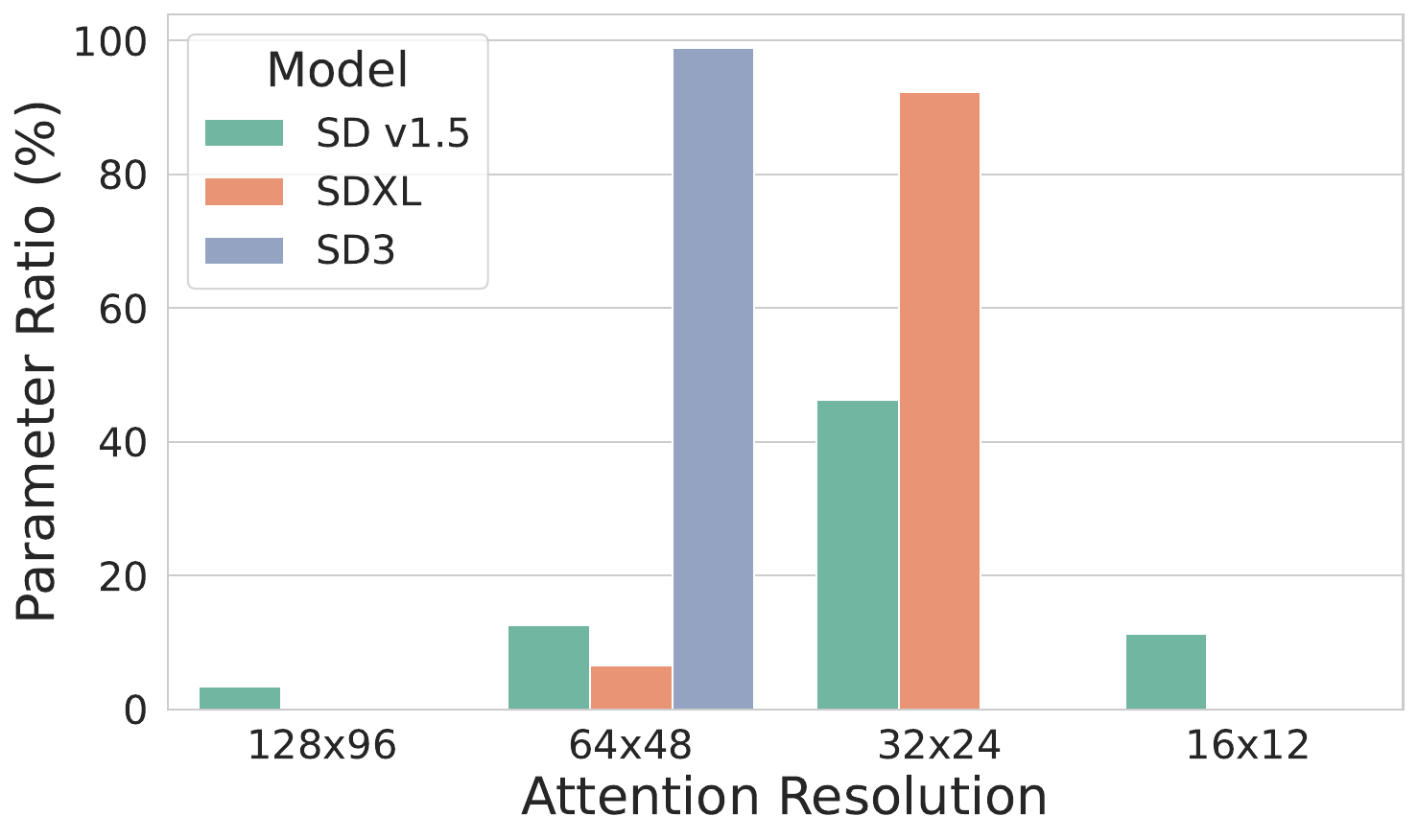}
    \vspace{-3mm}
    \caption{Attention-related parameter ratios at various resolutions.}
    \label{fig:advantage}
    \vspace{-5mm}
\end{figure}

\begin{figure*}[htb]
    \centering
    \includegraphics[width=0.89\textwidth]{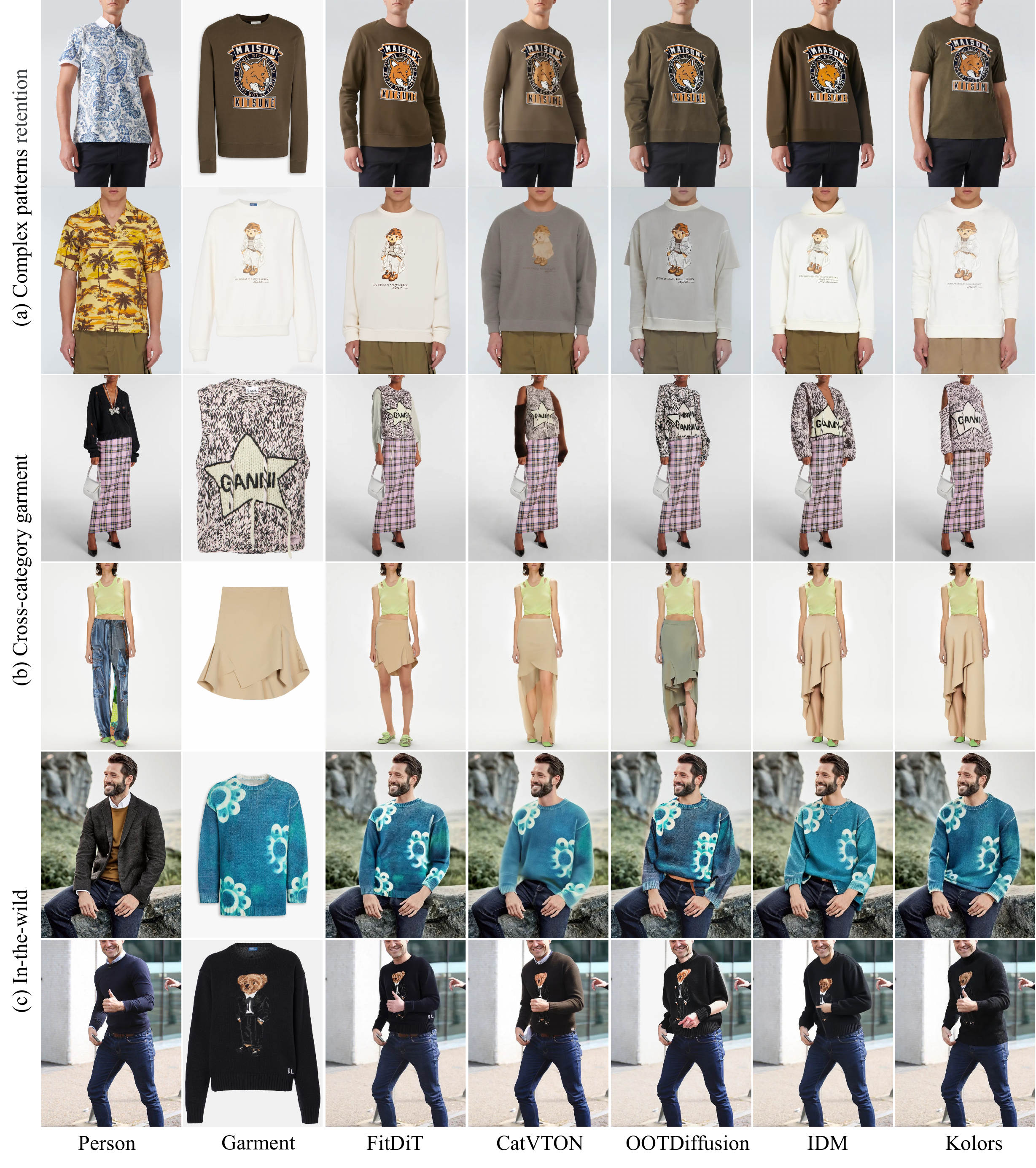}
    \vspace{-3mm}
    \caption{Visual results on CVDD with complex garment texture, cross-categories, and in-the-wild try-on. Best viewed when zoomed in.}
    \label{fig:main_fig}
    \vspace{-5mm}
\end{figure*}
\subsection{Advantages of using DiT for Try-on}
\label{AdvantagesDiT}
Existing try-on works are typically implemented based on Stable Diffusion v1.5~\cite{rombach2022high} or Stable Diffusion XL~\cite{podell2023sdxl}, both of which utilize a U-Net backbone architecture. 
In the context of the virtual try-on task, garment features from the reference branch are injected into the denoising branch using a hybrid attention mechanism. Consequently, the injection of high-resolution features is crucial for detail preservation. In Fig.~\ref{fig:advantage}, we illustrate the attention-related parameter ratios of different models at various latent resolutions, all using the same input resolution of ($1024 \times 768$). SDXL assigns 92\% of its parameters to the $32 \times 24$ resolution. For SD v1.5, the attention-related parameter ratios at latent resolutions higher than $64 \times 48$ only account 16\%. In contrast, SD3 allocates over 99\% of its parameters to the $64 \times 48$ resolution, providing more opportunities for high-resolution feature fusion. This makes DiT particularly suitable for tasks such as virtual try-on, which requires detail preservation.
It is clear that SD3 allocates over 99\% of its parameters to the $64 \times 48$ resolution, while SDXL assigns 92\% of its parameters to the $32 \times 24$ resolution. For SD v1.5, the attention-related parameter at latent resolution higher than $64 \times 48$ only accounts 16\%. 
Therefore, for tasks such as virtual try-on, which demand high fidelity in detail preservation, the DiT architecture emerges as a superior choice.

\begin{figure*}[htb]
    \centering
    \includegraphics[width=0.88\textwidth]{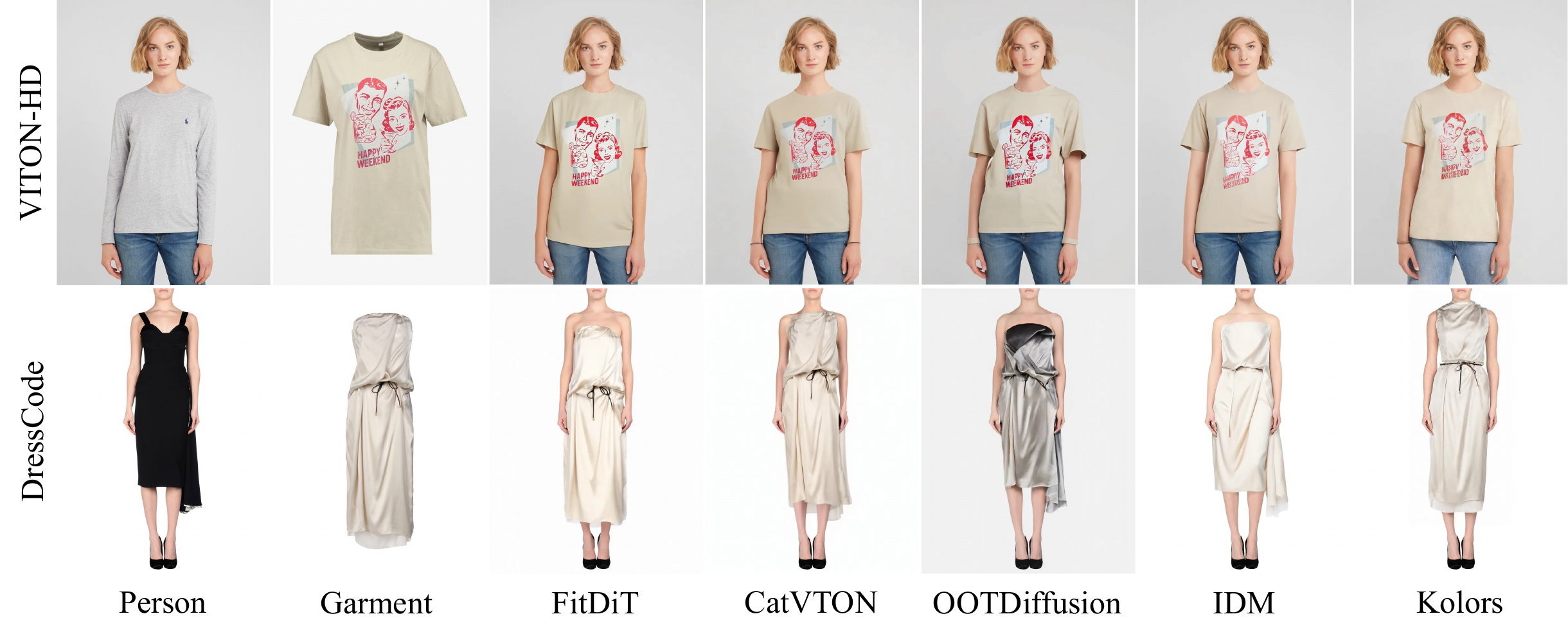}
    \vspace{-3mm}
    \caption{Visual results on DressCode and VTON-HD test set. Best viewed when zoomed in.}
    \label{fig:main_fig_public}
    \vspace{-3mm}
\end{figure*}

\begin{table*}[htbp]
    \centering
    \resizebox{\textwidth}{!}{%
    \begin{tabular}{lcccccccccccc}
        \toprule 
        \multirow{2}{*}{Methods} & \multicolumn{6}{c}{DressCode} & \multicolumn{6}{c}{VITON-HD} \\
        \cmidrule(lr){2-7} \cmidrule(lr){8-13}
        & \multicolumn{4}{c}{Paired} & \multicolumn{2}{c}{Unpaired} & \multicolumn{4}{c}{Paired} & \multicolumn{2}{c}{Unpaired} \\
        \cmidrule(lr){2-5} \cmidrule(lr){6-7} \cmidrule(lr){8-11} \cmidrule(lr){12-13}
        & SSIM $\uparrow$ & LPIPS $\downarrow$ & FID $\downarrow$ & KID $\downarrow$  & FID $\downarrow$ & KID $\downarrow$ & SSIM $\uparrow$ & LPIPS $\downarrow$  & FID $\downarrow$ & KID $\downarrow$ & FID $\downarrow$ & KID $\downarrow$ \\
        \midrule
        LaDI-VTON (2023) & 0.9025 &	0.0719&	4.8636&	1.5580&	6.8421&	2.3345	&	0.8763&	0.0911&	6.6044&	1.0672	&9.4095& 1.6866 \\
        StableVTON (2024) &-	&-&	-	&-	&-	&-		&0.8665	&0.0835	&6.8581&	1.2553&	9.5868	&1.4508 \\

        IDM-VTON (2024) &0.9228&	0.0478	&3.8001	&1.2012	&5.6159	&1.5536		&{0.8806}	&0.0789	&6.3381	& 1.3224	&9.6114	& 1.6387\\
        OOTDiffusion (2024) & 0.8975&	0.0725	&3.9497	&0.7198&	6.7019&	1.8630	&	0.8513&	0.0964&	6.5186&	{0.8961}	& 9.6733	&1.2061\\
        CatVTON (2024) & 0.9011 &	0.0705 & 3.2755 &0.6696 &5.4220 &1.5490   &
        {0.8694} & {0.0970} & {6.1394} & {0.9639} 
        & {9.1434} & {1.2666}\\
        \midrule
        {FitDiT (Ours)} & \textbf{0.9259} & \textbf{0.0431} & \textbf{2.6383} & \textbf{0.4990} & \textbf{4.7324} & \textbf{0.9011} & 
        \textbf{0.8985} & \textbf{0.0661} & \textbf{4.7309} & \textbf{0.1895} & \textbf{8.2042} & \textbf{0.3421} \\
        \bottomrule
    \end{tabular}
}
\vspace{-3mm}
    \caption{\small Quantitative results on VITON-HD and DressCode datasets. We compare the metrics under both paired (model's clothing is the same as the given cloth image) and unpaired settings (model's clothing differs) with other methods. 
    }
    \label{tab:main_results}
    \vspace{-3mm}
\end{table*}

\section{Experiments}
\subsection{Datasets and Implementation}
Our experiments are conducted on two publicly available fashion datasets: VITON-HD~\cite{choi2021viton} and DressCode~\cite{morelli2022dress}, along with a self-collected dataset called the Complex Virtual Dressing Dataset (CVDD), which consists of 516 high-resolution image pairs featuring complex garment textures and human poses. For quantitative comparison on two public datasets, we train two variants of models on the VITON-HD and DressCode separately using the official splits at a resolution of $1024 \times 768$. For quantitative comparison on CVDD, we first train a model with the combination of VITON-HD and DressCode at a resolution of $1024 \times 768$ and then fine-tune it at a higher resolution of $1536 \times 1152$ with only the parameters of the attention layers being trainable. This model is also utilized for qualitative comparisons. All experiments are conducted on 8 NVIDIA H20 GPUs, utilizing a total batch size of 32. We employ the AdamW optimizer~\cite{loshchilov2017fixing} with a learning rate of $ 3\times 10^{-5}$.

\subsection{Evaluation Metrics}
In paired try-on settings with ground truth in test datasets, we assess reconstruction accuracy using LPIPS~\cite{zhang2018unreasonable} and SSIM~\cite{wang2004image}, as well as the generation authenticity using FID~\cite{parmar2022aliased} and KID~\cite{binkowski2018demystifying}. In unpaired settings, we report only FID and KID due to the absence of ground truth.

\subsection{Qualitative Results}
For qualitative results, we compare our method with OOTDiffusion~\cite{xu2024ootdiffusion}, CatVTON~\cite{chong2024catvton}, IDM~\cite{choi2024improving} and Kolors-Virtual-Try-On~\cite{kolors}. The results are presented in Fig.~\ref{fig:main_fig} and Fig.~\ref{fig:main_fig_public}. Fig.~\ref{fig:main_fig_public} illustrates the results from VITON-HD and DressCode datasets, while Fig.~\ref{fig:main_fig} showcases results from our Complex Virtual Dressing Dataset (CVDD). Due to the superior design of the DiT architecture, which incorporates high-resolution feature injection, frequency loss, and garment priors evolution, FitDiT excels in retaining complex textures and small text, as demonstrated in Fig.~\ref{fig:main_fig}(a). This capability is crucial for real-world try-on applications. For cross-category try-on, FitDiT employs a dilated-relaxed mask strategy that effectively maintains the shape of the clothing during outfit changes, as shown in Fig.~\ref{fig:main_fig}(b). In contrast, other methods tend to fill the entire mask area, resulting in incorrect appearances. Additionally, we present virtual try-on in-the-wild, featuring complex backgrounds and human poses in Fig.~\ref{fig:main_fig}(c).

\subsection{Quantitative Results}
We conduct extensive experiments on VITON-HD, DressCode, and CVDD, as indicated in Tab.~\ref{tab:main_results} and Tab.~\ref{tab:fitdit_dataset}. FitDiT significantly outperforms all baselines across all datasets. Notably, on the CVDD, which features garments with complex textures and intricate patterns, FitDiT demonstrates substantial progress, highlighting the model's strengths in texture preservation.
As a significant performance milestone, FitDiT achieves a remarkable reduction in the KID error by \textbf{71.6\%} compared to the second-best method, OOTDiffusion, on the unpaired VITON-HD dataset.

\begin{table}[htbp]
    \centering
    \vspace{-1mm}
    \scriptsize
    \resizebox{0.9\linewidth}{!}{%
    \begin{tabular}{lcccccc}
        \toprule 
         \multirow{1}{*}{Methods} & \multicolumn{4}{c}{Paired} & \multicolumn{2}{c}{Unpaired} \\
         \cmidrule(lr){2-5} \cmidrule(lr){6-7} 
          & SSIM $\uparrow$ & LPIPS $\downarrow$ & FID $\downarrow$ & KID $\downarrow$  & FID $\downarrow$ & KID $\downarrow$ \\
        \midrule
        LaDI-VTON (2023) &  0.8431 &0.1432 & 26.4509 & 1.024 & 39.4821 & 3.0239
        \\
        IDM-VTON (2024)  &  0.8529 &0.1399 & 24.9510 & 0.7931 & 35.8422 & 1.1313
        \\
        OOTDiffusion (2024)  & 0.8397 &	0.1485 & 26.2757 & 1.1137  & 40.7213 &  4.3277 \\
        CatVTON (2024)  & 0.8457 & 0.1494 & 27.7435 & 1.7160 &  38.7899 & 3.4777\\
        \midrule
        {FitDiT (Ours)} & \textbf{0.8636} & \textbf{0.1130} & \textbf{20.7543} & \textbf{0.1602} & \textbf{33.4937} & \textbf{0.7434} \\
        \bottomrule
    \end{tabular}
}
    \vspace{-1mm}
    \caption{\small  Quantitative results on CVDD. 
    }
    \label{tab:fitdit_dataset}
    \vspace{-5mm}
\end{table}

\subsection{Ablation Study}
\begin{figure}[htb]
    \centering
    \includegraphics[width=0.88\linewidth]{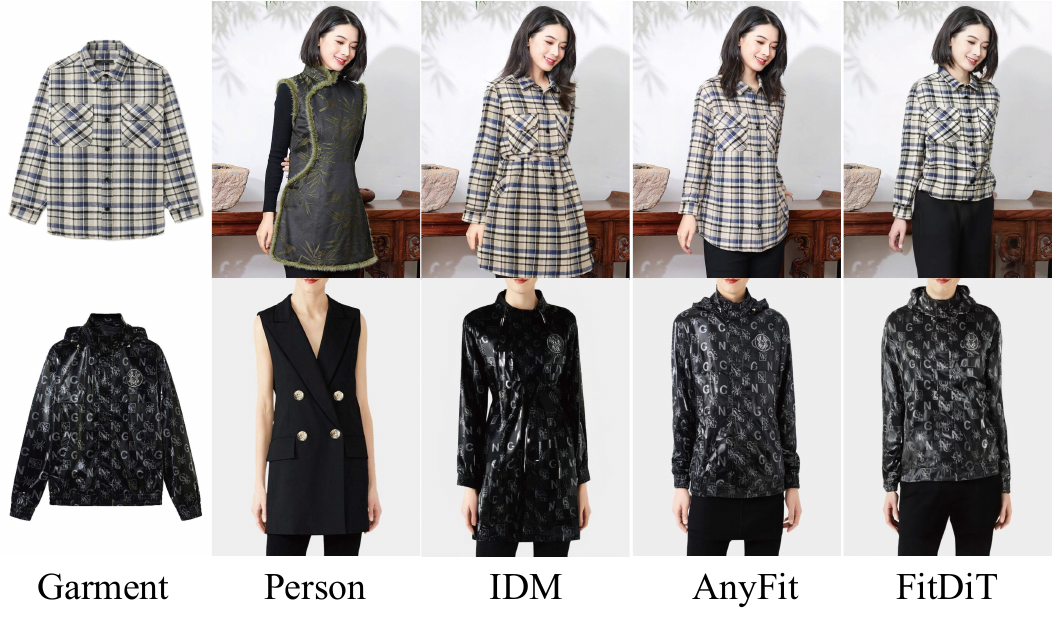}
    \vspace{-2mm}
    \caption{Visual validation of the role of dilated-relaxed mask.}
    \vspace{-4mm}
    \label{fig:aba_mask}
\end{figure}

\noindent\textbf{Dilated-relaxed mask.} We demonstrate the effectiveness of dilated-relaxed mask in Figs.~\ref{fig:relaxed_mask},~\ref{fig:main_fig} and~\ref{fig:aba_mask}. In the context of cross-category try-on, IDM, which employs a strict mask strategy, suffers from leakage of clothing shape, causing the model to fill the entire mask area. AnyFit, utilizing a parsing-free strategy, can somewhat determine the correct length of the garment. However, its masks are constructed solely based on keypoints, which is inadequate for generating appropriate agnostic masks. Given the significant length variations among different clothing styles, such as long down jackets and athletic vests, relying exclusively on keypoints may not suffice for generating suitable agnostic masks. As a result, the results of AnyFit remain slightly inferior to those of our FitDiT, which employs a dilated-relaxed mask strategy.

\begin{figure}[htb]
    \centering
    \includegraphics[width=0.98\linewidth]{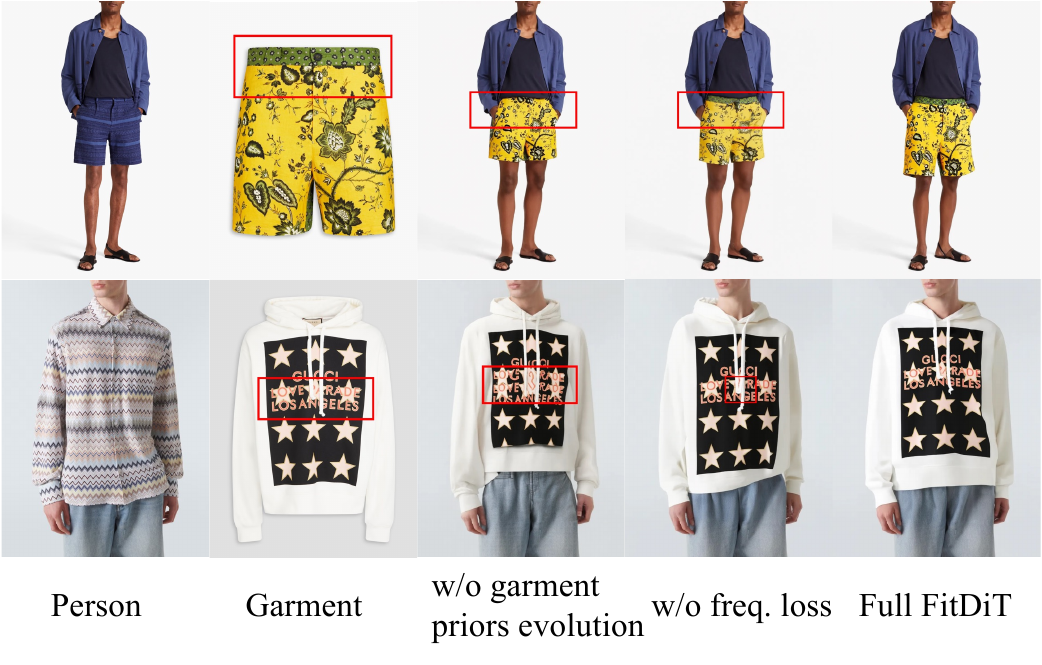}
    \vspace{-3mm}
    \caption{Visual validation of the role of garment priors evolution and frequency loss.}
    \label{fig:aba_2}
    \vspace{-1mm}
\end{figure}

\noindent\textbf{Frequency loss.} We demonstrate the effectiveness of frequency loss in Fig.~\ref{fig:aba_2} and Tab.~\ref{tab:aba}. By incorporating frequency loss during training, the resulting images exhibit sharper details and more accurately preserved textures (\textit{e.g.,} drawstring on the sweater).

\noindent\textbf{Garment priors evolution.} The effectiveness of garment priors evolution is illustrated in Fig.~\ref{fig:aba_2} and Tab.~\ref{tab:aba}. By fine-tuning the garment-dedicated extractor using garment-specific data, we achieve exceptional detail and richness. It effectively mitigates issues related to artifacts and texture loss in the generated results (\textit{e.g.,} font and pattern).

\begin{table}[htbp]
    \centering
    \scriptsize
    \resizebox{0.95\linewidth}{!}{%
    \begin{tabular}{lcccc}
        \toprule 
         Method & SSIM $\uparrow$ & LPIPS $\downarrow$ & FID $\downarrow$ & KID $\downarrow$ \\
        \midrule
        - w/o Frequency loss  &  0.8593 &  0.1239 & 22.6325  &  0.2960
        \\
        - w/o garment priors evolution   & 0.8578   &  0.1269 &  23.1786 &  0.5214  \\
        Full FitDiT  & \textbf{0.8636}  & \textbf{0.1130}  &  \textbf{20.7543} &  \textbf{0.1602} \\
        \bottomrule
    \end{tabular}
}
    \vspace{-1.5mm}
    \caption{\small  Ablation study results on CVDD. 
    }
    \label{tab:aba}
    \vspace{-5.5mm}
\end{table}

\subsection{Computational Analysis}
We analyze the model's computation cost in terms of inference time and GPU memory usage. All models operate with FP16 precision, and the input resolutions are set to $1024\times 768$, with 25 denoising steps. As shown in Tab.~\ref{tab:speed}, although FitDiT performs garment feature injection on high-resolution latent features, we customize the DiT structure specifically for the try-on task by removing the text encoder and text feature injection. This customization significantly reduces the model's computational complexity and parameter count. Consequently, FitDiT achieved an inference time of 4.57 seconds, which is 27\% faster than StableVITON and 54\% faster than IDM, while maintaining GPU memory usage consistent with IDM. Furthermore, by incorporating a sequential CPU offload technique, FitDiT can operate with less than 6GB of memory on a consumer GPU.

\begin{table}[htbp]
    \centering
    \scriptsize
    \resizebox{0.95\linewidth}{!}{%
    \begin{tabular}{lccccc}
        \toprule 
         Method & StableVITON & OOTDiffusion & IDM  & CatVTON & FitDiT \\
        \midrule
        Inference time (s) & 6.23 &  8.51 &  9.99 & 7.87  &  \textbf{4.57}
        \\
        GPU memory (M) & 10,978 & 8,962   &  19,504 &  \textbf{8,384} &  19,550  \\

        \bottomrule
    \end{tabular}
}
    \vspace{-1.5mm}
    \caption{\small  Computational analysis of different methods. 
    }
    \label{tab:speed}
    \vspace{-5.5mm}
\end{table}

\section{Conclusion}
We propose FitDiT to customize the Diffusion Transformer (DiT) structure specifically designed for virtual try-on via enhancing the high-resolution texture, and also provide insight analysis of DiT structure attention. Besides, our technical cornerstone lies in garment priors evolution, frequency domain learning, and dilated-relaxed mask augmentation, which effectively support texture-aware maintenance and size-aware fitting of garments.
Through extensive qualitative and quantitative evaluations, our research demonstrates superiority over state-of-the-art virtual try-on models, particularly in handling rich-textured garments and addressing cases of mismatched sizes. These findings mark a significant milestone in advancing the field of virtual try-on, paving the way for more intricate applications in real-world settings.

{
    \small
    \bibliographystyle{ieeenat_fullname}
    \bibliography{main}
}

\clearpage
\setcounter{page}{1}
\maketitlesupplementary

\section{More Qualitative Results  on  Texture-aware  Preservation}
\begin{figure*}[htb]
    \centering
    \includegraphics[width=0.98\textwidth]{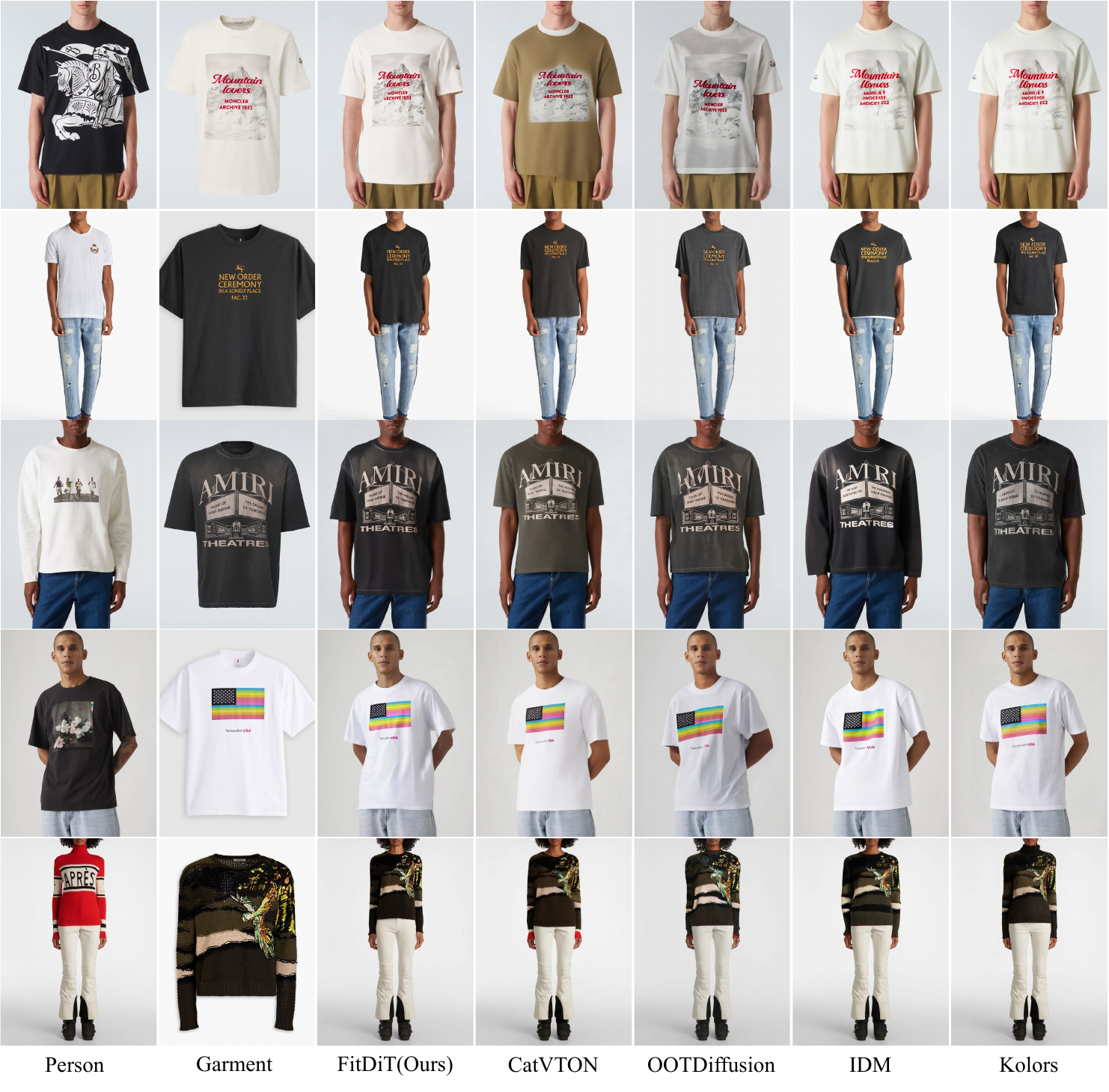}
    \vspace{-3mm}
    \caption{More qualitative comparisons with baselines on rich-texture preservation. Best viewed when zoomed in.}
    \label{fig:supp_detail}
    \vspace{-3mm}
\end{figure*}

As one of main challenges, texture-aware  preservation requires the model to well-capture intricate texture (\textit{e.g.,} pattern, font). Additional qualitative results of FitDiT on rich texture maintenance are shown in Fig. \ref{fig:supp_detail}.

\section{More Qualitative Results on Size-aware \\
Fitting}
\begin{figure*}[htb]
    \centering
    \includegraphics[width=0.98\textwidth]{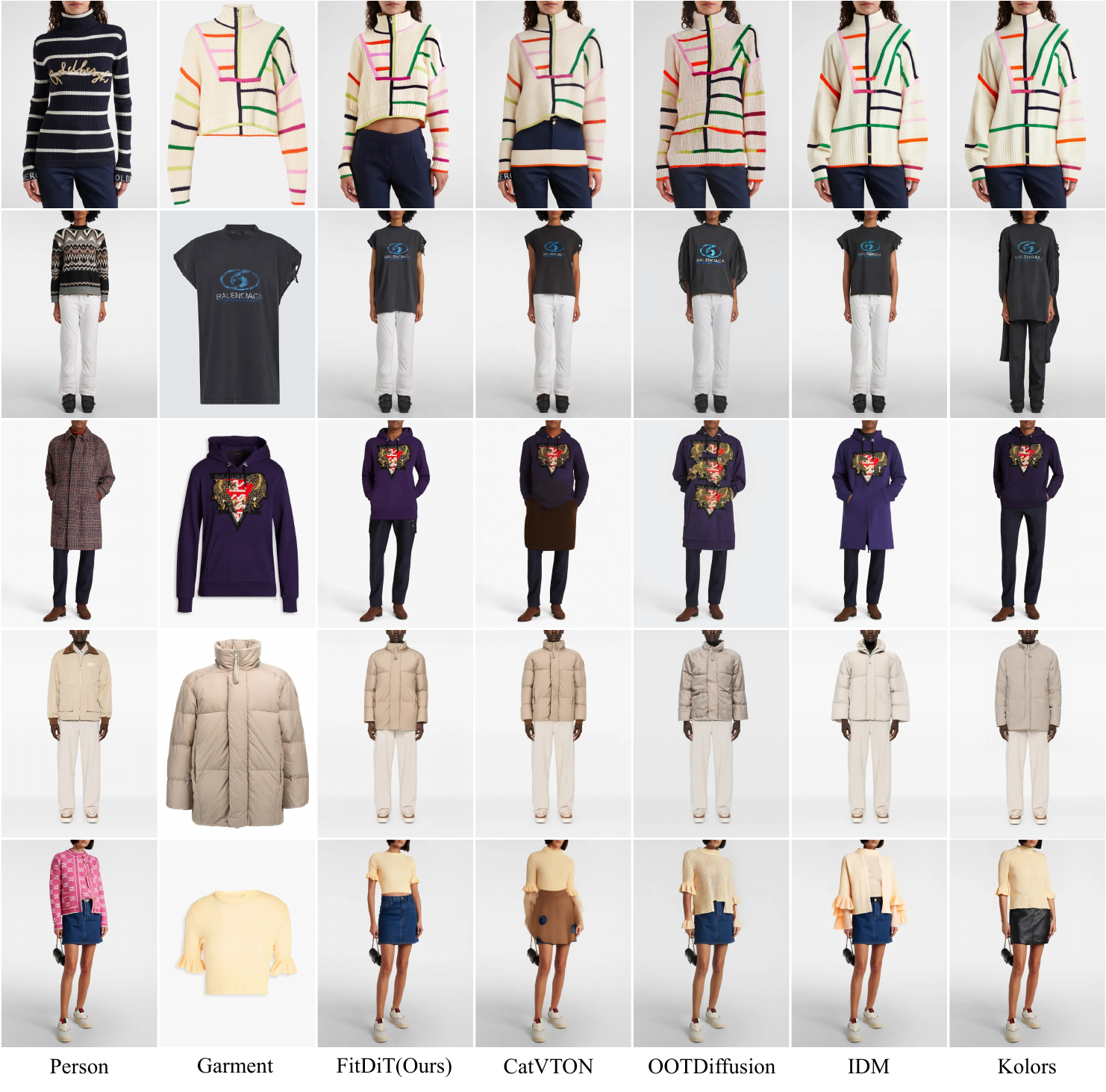}
    \vspace{-3mm}
    \caption{More qualitative comparisons with baselines on size-wise fitting. Best viewed when zoomed in.}
    \label{fig:supp_mask}
    \vspace{-3mm}
\end{figure*}

To verify the effectiveness of the  dilated-relaxed mask strategy on the size mismatching garments virtual try-on, we provide more qualitative results in Fig. \ref{fig:supp_mask}.

\section{More Qualitative Results in the Wild}
\begin{figure*}[htb]
    \centering
    \includegraphics[width=0.98\textwidth]{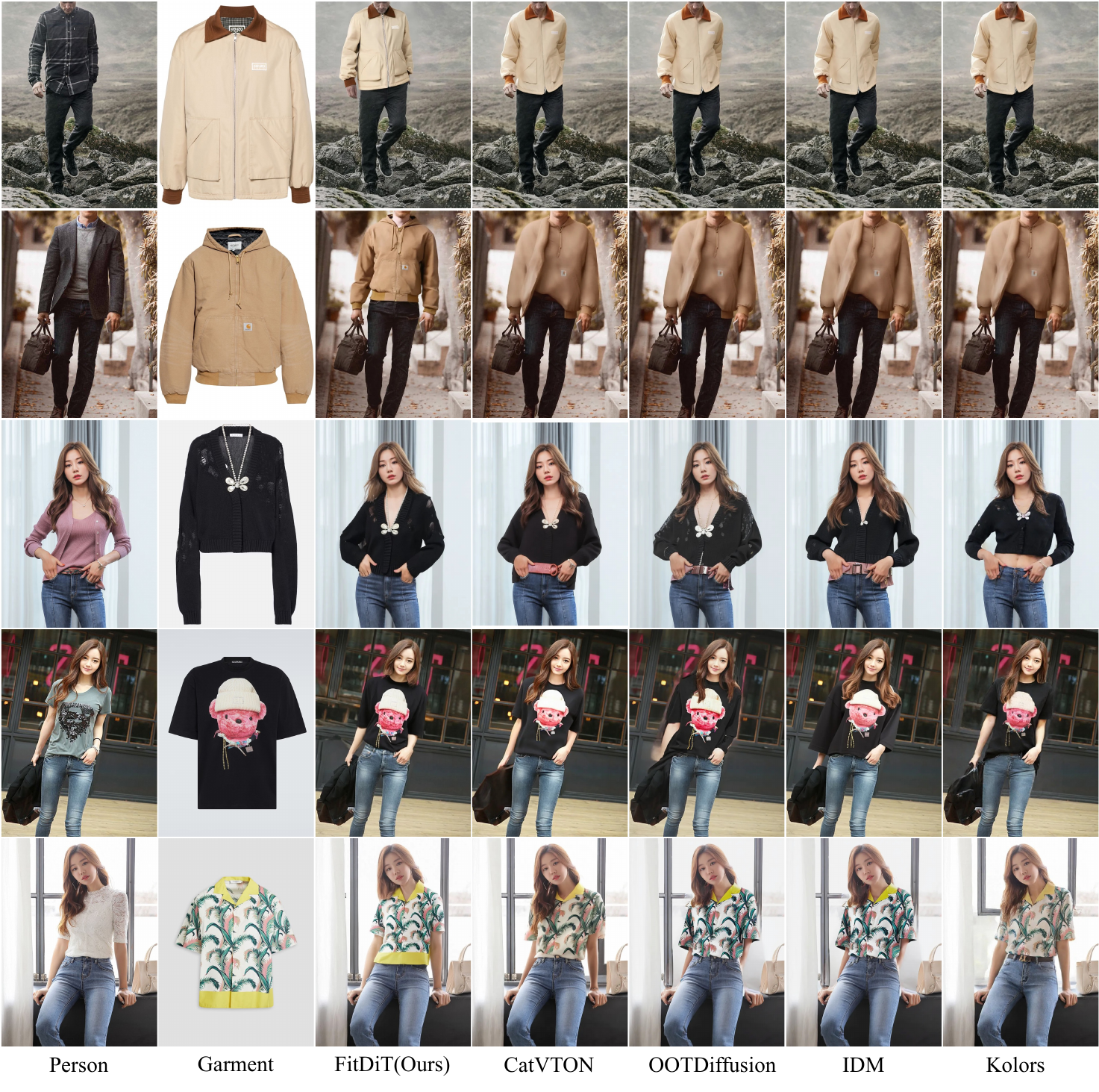}
    \vspace{-3mm}
    \caption{More qualitative comparisons with baselines in the wild. Best viewed when zoomed in.}
    \label{fig:supp_wild}
    \vspace{-3mm}
\end{figure*}

To examine the virtual try-on generalization ability of FitDiT in the wild, more qualitative results compared with other SOTA baselines are provided in Fig. \ref{fig:supp_wild}.

\section{More Qualitative Results on Complex Pose}
\begin{figure*}[htb]
    \centering
    \includegraphics[width=0.98\textwidth]{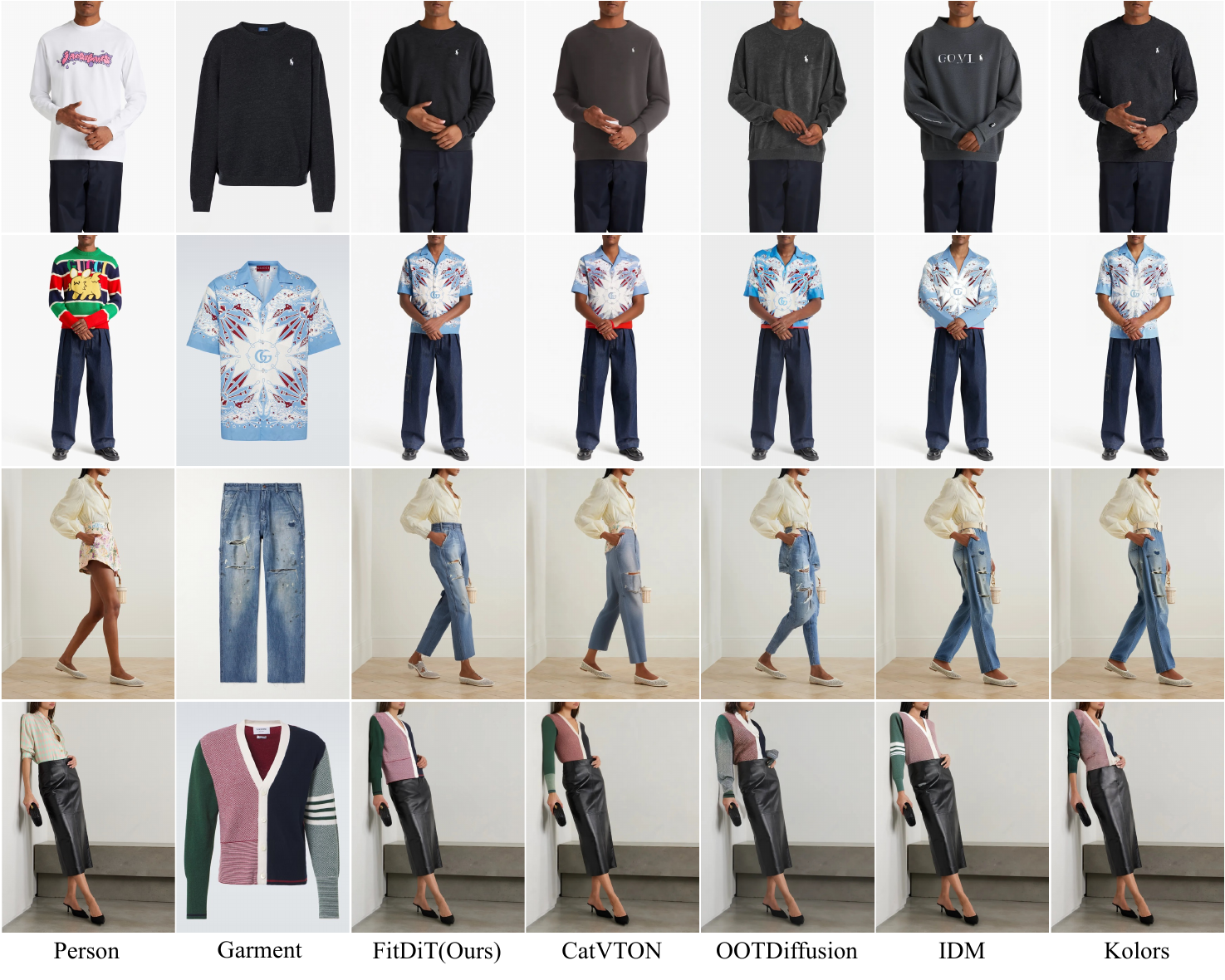}
    \vspace{-3mm}
    \caption{More qualitative comparisons with baselines on complex poses. Best viewed when zoomed in.}
    \label{fig:supp_pose}
    \vspace{-3mm}
\end{figure*}

The presence of complex poses beyond the training set poses a significant challenge in achieving satisfactory virtual try-on results. Fig. \ref{fig:supp_pose} demonstrates more qualitative comparisons with baselines on complex pose, which indicates the superiority of our FitDiT on complex poses.

\section{More Visual Results on Public Dataset}
\begin{figure*}[htb]
    \centering
    \includegraphics[width=0.98\textwidth]{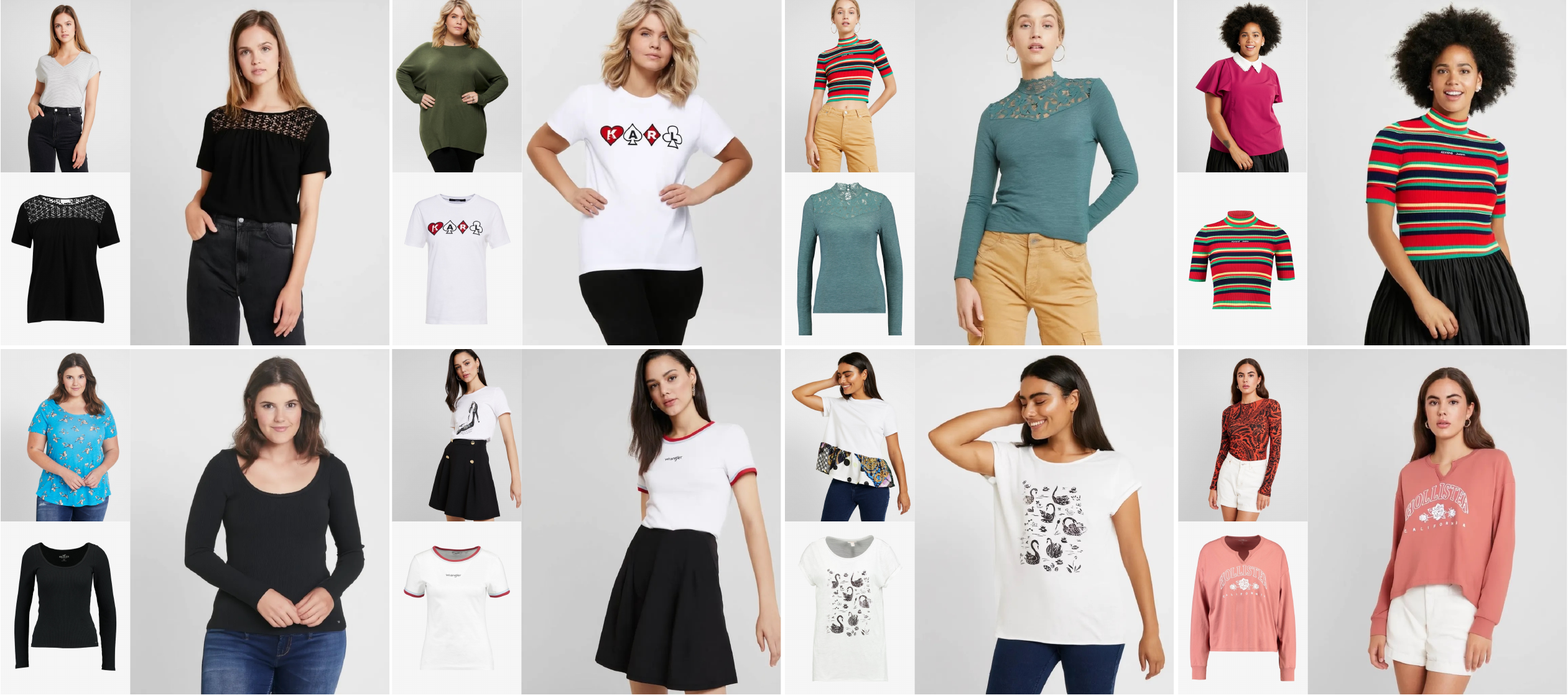}
    \vspace{-3mm}
    \caption{More visual results on the VITON-HD dataset. Best viewed when zoomed in.}
    \label{fig:supp_viton}
    \vspace{-3mm}
\end{figure*}

\noindent\textbf{VITON-HD dataset.}
More qualitative
results on the VITON-HD dataset are shown in Fig. \ref{fig:supp_viton}.

\begin{figure*}[htb]
    \centering
    \includegraphics[width=0.98\textwidth]{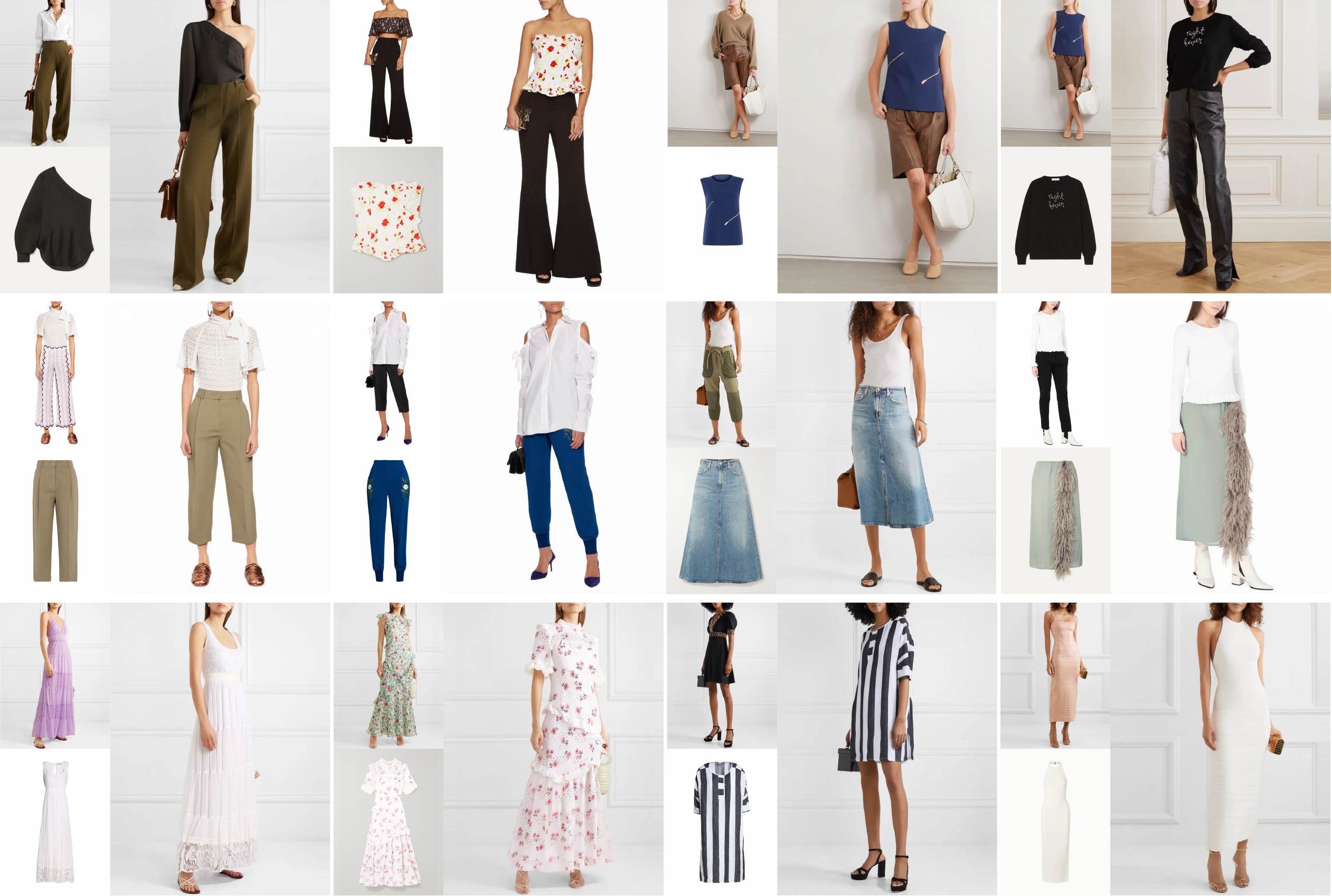}
    \vspace{-3mm}
    \caption{More visual results on the DressCode dataset. Best viewed when zoomed in.}
    \label{fig:supp_dresscode}
    \vspace{-3mm}
\end{figure*}

\noindent\textbf{DressCode dataset.}
Additional qualitative
results on the DressCode dataset are shown in Fig. \ref{fig:supp_dresscode}.

\section{Visual Cases of Complex Virtual 
Dressing  Dataset (CVDD)}
\begin{figure*}[htb]
    \centering
    \includegraphics[width=0.98\textwidth]{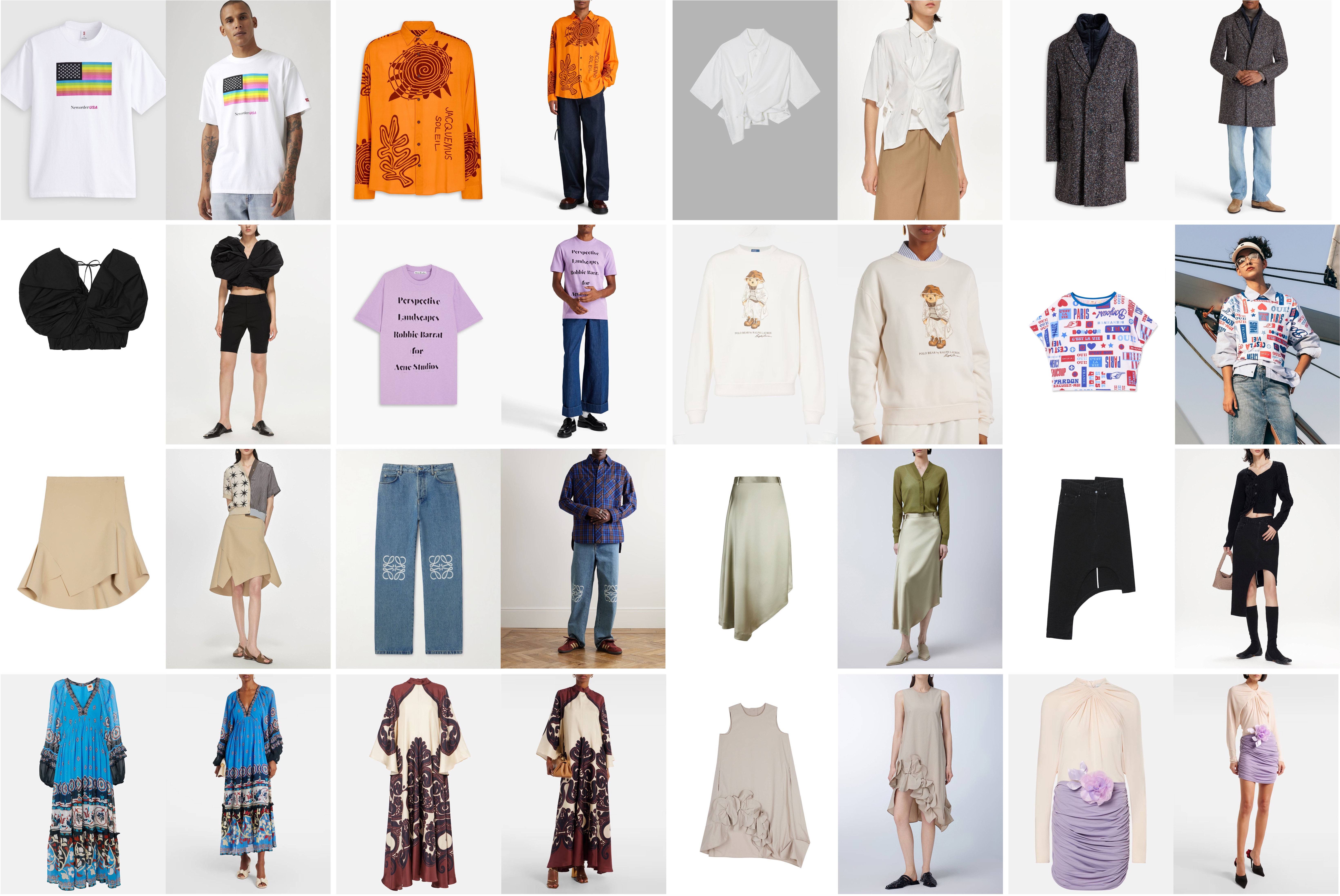}
    \vspace{-3mm}
    \caption{Examples from the CVDD dataset include virtual try-on images that feature complex textures, patterns, poses, and backgrounds. Best viewed when zoomed in.}
    \label{fig:supp_cvdd}
    \vspace{-3mm}
\end{figure*}

In Fig. \ref{fig:supp_cvdd}, we visualize more cases of CVDD to demonstrate the complexity including the intricate textures, diverse backgrounds, and complicated poses. 
We anticipate that the inclusion of self-collected challenging test cases will serve to validate the try-on model's capabilities in complex scenarios, thereby paving the way for its application in real-world scenarios with greater potential diversity.

\section{Applications on Model-to-model Try-on}

\begin{figure*}[htb]
    \centering
    \includegraphics[width=0.98\textwidth]{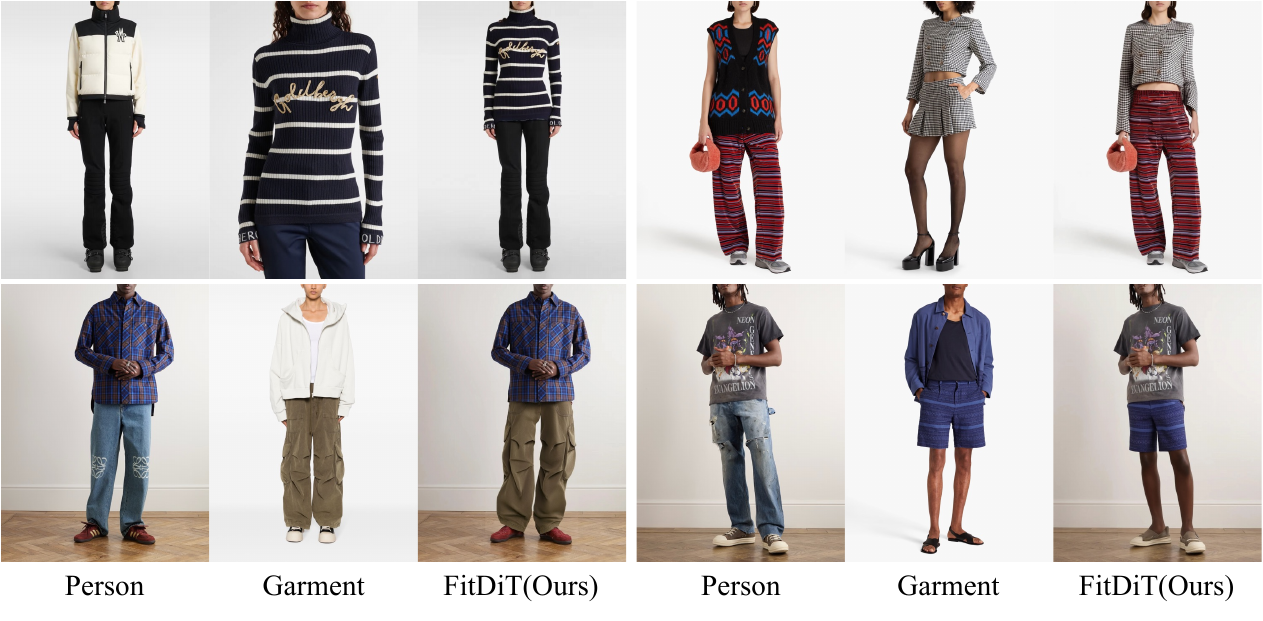}
    \vspace{-3mm}
    \caption{More visual results on the model-to-model virtual try-on applications. Best viewed when zoomed in.}
    \label{fig:supp_m2}
    \vspace{-3mm}
\end{figure*}

Fig. \ref{fig:supp_m2} illustrates FitDiT's capability to facilitate model-to-model virtual try-on applications, catering to the diverse demands encountered in real-world scenarios. In the model-to-model scenario, the target garment is sourced from another model rather than from a tiled clothing image. This is achieved by parsing the garment priors as inputs for the FitDiT try-on process.

\section{Limitation and Future Work}
Similar to other virtual try-on approaches, FitDiT faces challenges in accurately preserving the intricate details of human hands and fingers with complex pose, due to the lack of hand-relevant priors. In our future work, we intend to conduct a comprehensive investigation into hand priors, aiming to effectively preserve the realism of human fingers and achieve a heightened level of authenticity in virtual try-on applications.

\end{document}